\documentclass{article} 
\usepackage{iclr2027_conference,times}

\usepackage{amsmath,amsfonts,bm}

\def\eqref#1{equation~\ref{#1}}

\def\1{\bm{1}}

\DeclareMathAlphabet{\mathsfit}{\encodingdefault}{\sfdefault}{m}{sl}
\SetMathAlphabet{\mathsfit}{bold}{\encodingdefault}{\sfdefault}{bx}{n}

\usepackage{hyperref}
\usepackage{url}
\usepackage{inconsolata}
\usepackage[table]{xcolor}
\usepackage{colortbl}
\usepackage{graphicx}
\usepackage{tabularx}
\usepackage{color}
\usepackage{booktabs}
\usepackage{tablefootnote}
\usepackage{pifont} 
\usepackage{arydshln} 
\usepackage{multirow}
\usepackage{enumitem}
\usepackage{amsmath} 
\usepackage{amssymb}
\usepackage{inconsolata}
\usepackage{tcolorbox}
\usepackage{adjustbox}
\usepackage{makecell}
\usepackage{bbm}
\definecolor{hidden-draw}{RGB}{20,68,106}
\definecolor{hidden-pink}{RGB}{255,245,247}
\definecolor{maroon}{RGB}{148,78,99}
\definecolor{hidden-white}{RGB}{245,238,230}
\definecolor{hidden-yellow}{RGB}{255,248,227}
\definecolor{hidden-orange}{RGB}{243,215,202}
\definecolor{xm-purple}{RGB}{216, 218, 237}
\definecolor{xm-grey}{RGB}{242,242,242}
\newtcolorbox[list inside=prompt,auto counter]{prompt}[1][]{
    colbacktitle=xm-purple!90,
    colback =xm-grey!30,
    coltitle=black,
    fontupper=\footnotesize,
    boxsep=5pt,
    left=0pt,
    right=0pt,
    top=0pt,
    bottom=0pt,
    boxrule=0.5pt,
    #1,
}
\usepackage{booktabs}
\definecolor{websectiongray}{RGB}{242,242,242}
\definecolor{weboursblue}{RGB}{228,238,250}
\definecolor{websoftgreen}{RGB}{224,245,230}
\definecolor{websoftyellow}{RGB}{255,246,215}
\definecolor{webposgreen}{RGB}{28,125,70}
\definecolor{webnegred}{RGB}{190,65,55}
\definecolor{green}{RGB}{239, 251, 238}

\title{Retrospective Progress-Aware Self-Refinement for LLM Agent Training}

\author{Xinbei Ma$^{1,3}$, Congmin Zheng$^{1}$, Jiyang Qiu$^{1}$, Jiale Hong$^{1}$, Yao Yao$^{1}$,\\ \textbf{Xiangmou Qu$^{2}$, Jiaxin Yin$^{2}$, Xingyu Lou$^{2}$, Jun Wang$^{2}$, Weiwen Liu$^{1}$, Weinan Zhang$^{1}$,} \\ \textbf{Zhuosheng Zhang$^{1}$, Hai Zhao$^{1}$, Bryan Hooi$^{3}$}\\
\textsuperscript{1}Shanghai Jiao Tong University, 
\textsuperscript{2}OPPO Research Institute,
\textsuperscript{3}National University of Singapore
}

\iclrfinalcopy 
\begin{document}

\maketitle



\begin{abstract}
Long-horizon LLM-based agents receive rich environmental observations during interaction, yet outcome rewards provide limited explicit supervision about how individual actions advance task completion.
We investigate whether agents can turn this interaction evidence into useful training signals through retrospective progress assessment.
A WebShop pilot shows that direct progress prompting reduces task success, whereas hindsight-annotated demonstrations improve it.
We introduce \textbf{RePro}, \textbf{Re}trospective \textbf{Pro}gress-Aware Training, with a \textit{forward-then-reflect} rollout: the agent estimates progress while acting, then reassesses each step using the completed trajectory and outcome.
After warmup with externally generated demonstrations, policy optimization combines self-generated progress differences, online-retrospective alignment, and format rewards with environment feedback, requiring neither a separate process reward model nor ongoing teacher annotation.
Experiments on WebShop, ALFWorld, and Sokoban show that  RePro enhances the Qwen family's performance, with up to $11.57\%$ success rate gains.

\end{abstract}

\section{Introduction}

Reinforcement learning with verifiable rewards (RLVR) has enabled Large language models (LLMs) to learn from experience through outcome rewards, substantially improving the reasoning capabilities \cite{deepseekmath, feng2026groupingroup, dong2025arpo}.
These rewards, however, provide no explicit supervision about progress at intermediate steps. 
In multi-turn agentic tasks, agents also receive observations after taking actions, accumulating rich environmental feedback about how their interactions affect the environment \cite{xie2024osworld, liu2026scalecua, shridhar2020alfworld, wang2026enact}. 
Turning this rich feedback into useful training signals requires assessing how each interaction with the environment associates with the task goal.


Human metacognition offers a perspective on deriving training signals from these interactions: when pursuing a goal, people monitor what has been accomplished, what remains, and whether they are on track \citep{lin2026plan, li2025selfbudgeter, han-etal-2025-token}.
Progress assessment makes this perspective concrete by evaluating how intermediate interactions advance task completion.
Unlike a standard scalar value estimate of expected return, each assessment includes a verbalized rationale that draws on the LLM's semantic knowledge to explain how the observed state relates to the task goal.
Such assessments can be generated through the actor policy's own retrospective reflection, providing a source of process-level supervision without a separately trained process reward model.
This motivates us to ask whether agents can generate useful progress assessments from their interaction histories and use them as training signals to complement sparse outcome rewards.


In a preliminary study on WebShop, we investigate whether agents can generate progress assessments that improve decisions.
We compare two ways of eliciting progress assessments: directly prompting the agent to assess its current state, and teaching the agent with demonstrations of progress annotated retrospectively from completed trajectories with known outcomes.
In both settings, the resulting assessments inform subsequent
action selection.
Relative to a baseline without progress assessments, direct prompting lowers task success by 5.20 on average, whereas hindsight-guided prompting raises it by 3.35 points.
This contrast suggests that useful progress assessment does not emerge simply from asking agents to evaluate their current state, but can be supported by retrospective information.
These findings motivate leveraging progress assessment from retrospective trajectory outcomes.

This paper proposes \textbf{RePro} (\textbf{Re}trospective \textbf{Pro}gress-Aware Training), a two-phase framework that uses the agent’s own retrospective assessments to construct step-wise training signals to shape policy rewards, while an auxiliary alignment objective trains the progress predictions produced during execution toward the retrospective estimates.
We augment trajectories with a \textbf{forward-then-reflect} paradigm: the agent first \emph{forward} to execute the task while generating online progress estimates; once completing the task and knowing the outcome, the agent retrospectively reassesses its step-wise progress anchored by the final result.
Based on this, RePro first applies a stage of \textbf{Retrospection Warmup}: a small set of external-LLM demonstrations teaches the agent the forward-then-reflect format for retrospective reflection.
After that, \textbf{RePro-PO} leverages the retrospective progress signal by a composite reward, comprising retrospective progress shaping, online-retrospective alignment, and format regularization.

We evaluate RePro across LLMs of varying sizes on WebShop, ALFWorld, and Sokoban.
On WebShop, RePro improves the absolute task success rate over baseline training by $+8.98\%$, $+11.57\%$, and $+5.82\%$ across different model sizes.
It also yields consistent gains on ALFWorld and Sokoban, and outperforms all ablation variants.
Progress-quality analysis further provides behavioral evidence for improved progress awareness: compared with the prompting baseline, RePro's intermediate estimates provide substantially stronger discrimination between trajectories that eventually succeed and those that fail.
Our contributions are threefold:
\begin{enumerate}[leftmargin=*,itemsep=1pt,topsep=1pt]
    \item We show that directly prompting agents to estimate progress can hurt task performance, whereas retrospective demonstrations improve it, motivating the use of hindsight for progress assessment.
    \item We introduce RePro, which uses the agent's own retrospective reasoning over completed trajectories and outcomes to generate step-wise progress assessments and incorporate them into policy training.
    \item Across three agentic benchmarks and multiple model scales, we demonstrate RePro's effectiveness through performance comparisons and ablations, and characterize its retrospective progress assessments through behavioral analysis.
\end{enumerate}
\section{Related Work}

\subsection{LLM Agent Training}
LLM-based agent training has long drawn inspiration from reinforcement learning, showing that agent behavior can be improved by reusing past interaction trajectories as training signals \citep{mnih2015human,schulman2017proximal,xiong2024watch, song2024trial, cao2025pgpo}.
Recently, critic-free policy optimization has improved efficiency by estimating advantages from rollout groups, trajectory-level comparisons, or improved sampling strategies~\citep{guo2025deepseek,deepseekmath,feng2026groupingroup,dong2025arpo,dong2025aepo,zheng2025group}, and 
have been widely applied to LLM agent training across diverse environments \citep{qian2026toolrl,xi2026agentgymrl}.

Reward design remains a central challenge in long-horizon agent RL.
Outcome-based methods suffer from sparse terminal feedback~\citep{guo2025deepseek,deepseekmath}; process supervision introduces expensive annotations, external reward models, or engineered subgoals~\citep{lightman2024lets,zou2026reasonfluxprm,xi2026agentprm,zheng2026adaptive,wang2026subgoal} or relies on environment-specific signals~\citep{lu2026ui}.
Recent work has also revisited \textit{critics}, combining a critic or value model as compensation \citep {li2026turn, hou2026single}.
Alternatively, existing progress-based methods explicitly model task-completion progress through heuristic estimates or supervision from stronger LLMs~\citep{zhang2025progrm,zhang2026progress,chai2025parl}.
Differently, our method uses minimal external supervision and learns progress awareness from the agent's own completed trajectory outcomes, internalizing progress estimation into the agent itself without relying on an additional reward model, and at test time, jointly estimates its progress, helping make decisions.

\subsection{Self-Improving LLMs via Reflection}
Reflection improves agent behavior by turning past experience into reusable guidance. 
Failed episodes are converted into verbal feedback or episodic memory \citep{shinn2023reflexion,qu2024recursive}, while others ground reflection in execution states or evolving contextual playbooks \citep{kim2025reflact,guo2026towards,cai2026a,zhang2025agentic}. 
Recent work further extends reflection to uncertainty-aware test-time adaptation, reusable experiential heuristics, and pre-execution plan critique \citep{acikgoz2025self,ge2025samule,allard2026experiential,wang2026preflect}.
These methods can elicit reflective behaviors already present in the base model, but do not directly internalize new reflection capabilities into the agent.

Training-time methods instead aim to internalize reflection through learning. 
They improve meta-introspection or reward useful self-reflective tokens \citep{li2025reflectevo,bensal2025reflect}, support staged self-correction and consistency \citep{kumartraining,zhang2026consistent}, and enable self-evolution \citep{ou2026serl}.
Our work follows this line in internalizing self-improvement signals into the agent itself.
Rather than correcting individual actions, our method introduces retrospective, progress-aware reflection, in which the agent uses completed trajectories to assess task-level progress at intermediate steps.
This yields self-generated progress targets that train the agent to track \textit{where it stands} in a long-horizon task, without relying on an additional reward model.
Detailed comparisons with related methods are provided in Appendix~\ref{app:positioning}.

\section{Preliminaries}
\subsection{Formulation}
Given a natural language goal $g$, an LLM-based agent interacts with an environment to complete the task. This interaction is modeled as a Partially Observable Markov Decision Process (POMDP)~\citep{kaelbling1998planning}, defined by the tuple $(\mathcal{S}, \mathcal{A}, \mathcal{O}, \mathcal{P}, \mathcal{R})$, where $\mathcal{S}$, $\mathcal{A}$, $\mathcal{O}$, $\mathcal{P}$, and $\mathcal{R}$ denote the state space, action space, observation space, transition kernel, and reward function.

At each step $t$, the agent receives a partial observation $o_t \in \mathcal{O}$ of the underlying state $s_t \in \mathcal{S}$, and samples an action from its policy conditioned on the interaction history $h_t = (g, o_1, a_1, \dots, o_t)$:
\begin{equation}
\setlength{\abovedisplayskip}{5pt}
\setlength{\belowdisplayskip}{5pt}
a_t \sim \pi_\theta(\cdot \mid h_t).
\end{equation}
The environment transitions according to $\mathcal{P}(s_{t+1} \mid s_t, a_t)$, producing observation $o_{t+1}$.
This sequential interaction forms a trajectory
\begin{equation}
\setlength{\abovedisplayskip}{5pt}
\setlength{\belowdisplayskip}{5pt}
\tau = \{g,\, [(o_t, a_t)]_{t=1}^{T}\},
\label{eq:trajectory}
\end{equation}
which terminates upon an environment signal or agent decision at step $T$, yielding a task reward $r = \mathcal{R}(\tau)$.
The policy $\pi_\theta$ is optimized via RL to maximize the expected task reward.

Under partial observability, the agent has access only to a bounded window of its interaction history and partial observations of the environment, while its intermediate task progress remains latent and feedback is typically limited to a sparse terminal reward.
This motivates our study to infer and verbalize progress awareness and convert it into progress-aware reward signals.

\subsection{Pilot study}
\label{sec:pilot}
Under the formulation above, the agent receives an environmental observation after each action. These observations provide step-wise evidence of how the task is unfolding, which the agent may use to estimate its current progress $p_t$ and inform subsequent actions $a_{\ge t}$.

Our first question is: \textbf{can progress assessment help an agent act?} 
An assessment of how far the task has progressed may inform the agent's next decision. 
We study this question on WebShop\cite{yao2022webshop} by placing a progress-assessment step before action prediction at each interaction turn and measuring the resulting task success rate. 

The following question is: \textbf{how to elicit the progress assessment.}
At each decision-making time, the agent must assess progress from an incomplete trajectory; completed trajectories and their outcomes, however, provide additional context for progress assessment.
Therefore, we compare two methods:
\textbf{(1) direct assessment}, where the agent estimates progress from its current interaction history; and
\textbf{(2) retrospectively guided assessment}, 
where we collect completed trajectories and their outcomes from a separate training set. The same agent model retrospectively annotates the progress at each observation using the full traces and outcomes; at test time, these annotated trajectories serve as in-context demonstrations guiding the progress assessment.

In both cases, the agent first estimates its progress upon receiving observations and then appends that assessment to the action-prediction prompt for a separate call without demonstrations.
To ablate prompts and added contexts, we include a random baseline by replacing the model's online progress estimates or demonstrations with randomly sampled percentage values and fabricated rationales. Details are provided in Appendix~\ref{app:pilot_details}.

\begin{figure}[htb]
    \centering
    \includegraphics[width=0.95\linewidth]{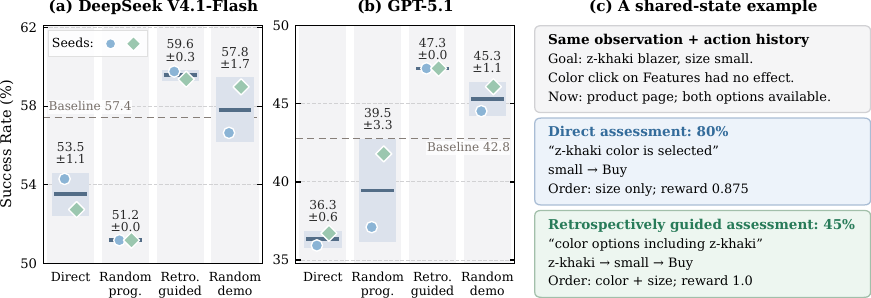}
     \caption{Pilot study results on WebShop. Online progress prompting hurts performance, while retrospective demonstrations for progress improve it.}
    \label{fig:Pilot}
\end{figure}

Figure~\ref{fig:Pilot} shows that the utility of progress assessment depends on how it is elicited.
(1) Direct assessment lowers mean success rates on both models (DeepSeek: 57.4\%$\to$53.5\%; GPT-5.1: 42.8\%$\to$36.3\%), while Random Progress also underperforms the baselines.
(2) Retrospectively guided assessment instead achieves 59.6\% and 47.3\%, improving over the baselines by an average of 3.3 percentage points.
Random Demonstration preserves the original observation sequences but replaces progress annotations with random numbers and fabricated rationales.
It achieves 57.8\% and 45.3\%, below retrospectively guided assessment on both models.
Although we do not establish the accuracy of the progress estimates, outcomes suggest that informative progress annotations provide benefits beyond trajectory content, while direct prompting does not reliably elicit assessments that improve decision-making.

(3) Figure~\ref{fig:Pilot}(c) illustrates a concrete difference: given the same action-observation history, direct assessment incorrectly treats a color as already selected with a 80\% procress percentage, whereas retrospectively guided assessment describes it as available with a 45\% completion.
This branch leads to different outcome of subsequent purchases.
Together, these observations motivate learning progress assessment from retrospective trajectory outcomes, rather than relying on direct prompting to elicit useful self-assessment.

\begin{figure*}[t]
    \centering
    \vspace{-10pt}
    \includegraphics[width=0.9\textwidth]{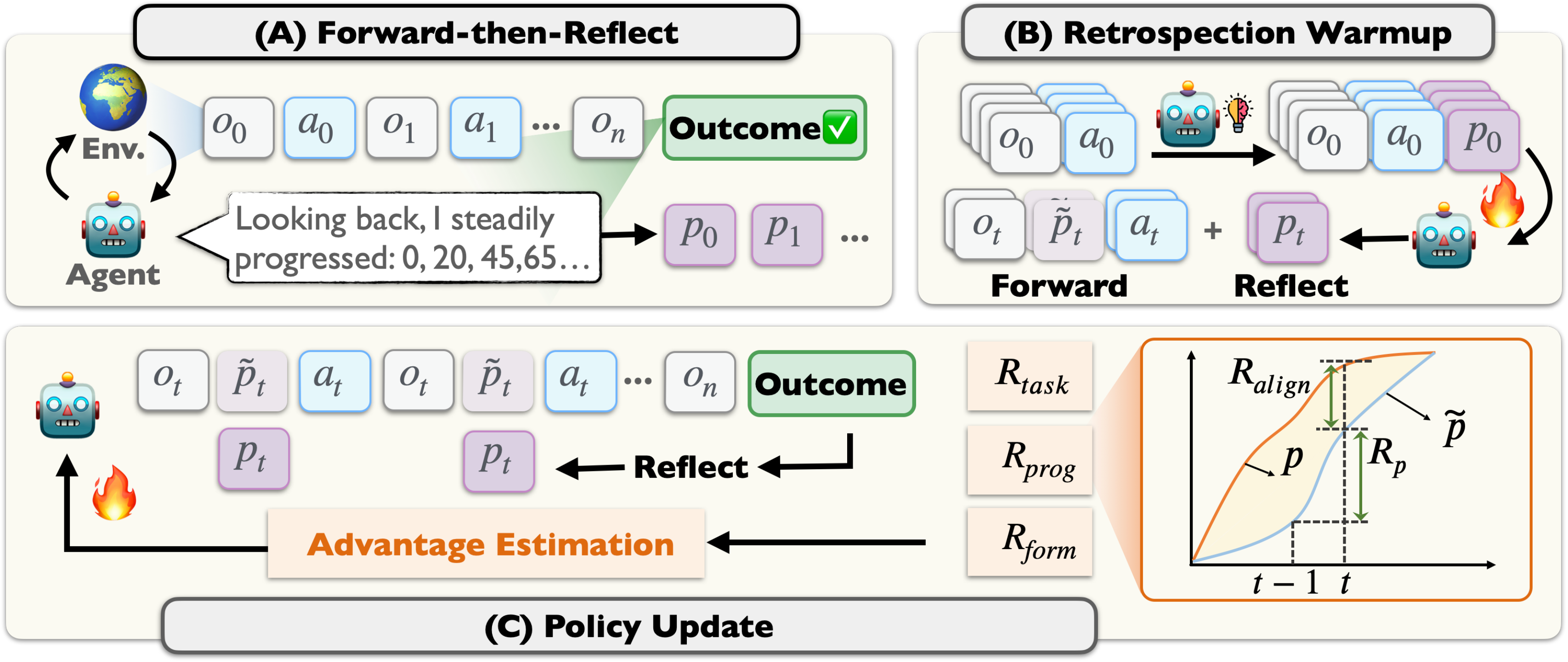}
    \vspace{-5pt}
    \caption{Overview of \textbf{RePro}. (A)\textbf{Forward-then-reflect}: the agent executes actions with online progress estimates, then retrospectively reassesses progress after observing the final outcome.
(B) \textbf{Retrospection Warmup} to initialize retrospective reflection behavior.
(C) \textbf{Policy Update} integrates retrospective progress into RL reward.}
    \label{fig:framework}
    \vspace{-10pt}
\end{figure*}

\section{Retrospective Progress-Aware Training}
Our preliminaries motivate us to use completed trajectories to generate progress assessments for training. We propose \textbf{RePro} (\textbf{Re}trospective \textbf{Pro}gress-Aware Training), shown in Figure~\ref{fig:framework}, where the agent reassesses its own trajectory after each episode and uses the resulting step-wise progress signals in policy optimization. RePro first uses \textbf{Retrospection Warmup} (\S\ref{sec:retrospective}) to initialize this reflection behavior, followed by \textbf{RePro-PO} (\S\ref{sec:rl-calibration}) to train with the its own retrospective assessments.

\subsection{Forward-then-Reflect}
By adding retrospective reflection after each episode, RePro turns the agent’s working pattern into a forward-then-reflect process. 

\noindent \textbf{Forward Execution.} At each step $t$, the agent receives the partially observed interaction history $h_t$,  generates a reasoning trace, and predicts the progress assessment and an action :
\begin{equation}
\setlength{\abovedisplayskip}{5pt}
\setlength{\belowdisplayskip}{5pt}
\begin{split}
(\tilde{p}_t, a_t) \sim \pi_\theta(\cdot \mid h_t),
\end{split}
\label{eq:forward-execution}
\end{equation}
where $\tilde{p}_t \in [0,1]$ denotes the agent’s verbalized online estimate of task completion at each decision point $t$ after observing $o_t$.
They are verbalized as percentages and normalized to $[0,1]$. The reasoning trace is omitted from the notation.
$a_t$ is the action to execute in the environment, yielding the next observation $o_{t+1}$.
The trajectory continues until the terminal step $T$, returning task reward $r$.

\noindent \textbf{Retrospective Reflection.} 
After the episode, the agent reviews the complete trajectory $\tau$ and outcome $r$. Prompted to reason about how the task progressed, it produces retrospective estimates for the previously acted steps.
We denote this prompted mode as $\pi_\theta^{\text{retro}}$:
\begin{equation}
\setlength{\abovedisplayskip}{5pt}
\setlength{\belowdisplayskip}{5pt}
\begin{split}
(p_1, \dots, p_T) \sim \pi_\theta^{\text{retro}}(\cdot \mid \tau, r),
\end{split}
\label{eq:retrospective-progress}
\end{equation}
Here, $p_t \in [0,1]$ retrospectively assesses task completion
after observing $o_t$ and before taking $a_t$, which are verbalized as percentages and normalized to $[0,1]$. 
The trajectory is then re-organized by associating the retrospective estimates $p_t$ with their corresponding steps:
\begin{equation}
\setlength{\abovedisplayskip}{5pt}
\setlength{\belowdisplayskip}{5pt}
\tau_{\text{retro}} = \{g, [(o_t, a_t, p_t, \tilde{p}_t)]_{t=1}^T\}.
\label{eq:retrospective-trajectory}
\end{equation}
The known outcome anchors the terminal assessment, and the complete trajectory provides context for assessing the intermediate steps.
  
\subsection{Retrospection Warmup}
\label{sec:retrospective}
Outcome-based RL trains the agent to act in the environment, but does not by itself establish the behavior of reflecting on a completed trajectory when prompted. We therefore use supervised warmup to initialize retrospective reflection.
We collect successful rollout trajectories and use an external model $\pi_{\text{demo}}$ (DeepSeek-V4) to generate each compact record with a retrospective reflection. 
Then, we replace each online progress estimate in the original trajectory with its retrospective counterpart and append the complete reflection as a final turn, yielding the warmup dataset:
\begin{equation}
\begin{aligned}
\setlength{\abovedisplayskip}{7pt}
\setlength{\belowdisplayskip}{7pt}
&\mathbf{p}_{\tau} = (p_1,\dots,p_T)
\sim \pi_{\text{demo}}^{\text{retro}}(\cdot\mid\tau,r),
\quad \tau\in\mathcal{D}_{\text{succ}}, \\
&\mathcal{D}_{\text{warmup}} = \{(\tau,\mathbf{p}_{\tau})
\mid \tau\in\mathcal{D}_{\text{succ}}\}.
\end{aligned}
\end{equation}

The agent is fine-tuned on this dataset using a cross-entropy loss to generate actions and progress estimates at each step, followed by a retrospective reflection after the episode.


\begin{equation}
\setlength{\abovedisplayskip}{7pt}
\setlength{\belowdisplayskip}{7pt}
\mathcal{L}_{\text{SFT}}
= -\mathbb{E}_{(\tau,\mathbf{p}_{\tau})\sim\mathcal{D}_{\text{warmup}}}
\left[
\sum_{t=1}^{T}\log\pi_\theta(p_t,a_t\mid h_t)
+\log\pi_\theta^{\text{retro}}(\mathbf{p}_{\tau}\mid\tau,r)
\right].
\end{equation}

\subsection{RePro-PO: Policy Optimization with Retrospective Progress}
\label{sec:rl-calibration}
The reward integrates the outcome task performance with progress-specific objectives.
\begin{equation}
\setlength{\abovedisplayskip}{5pt}
\setlength{\belowdisplayskip}{5pt}
R(\tau)
= \sum_{t=1}^{T} r_t + R_{\text{boundary}}(\tau),
\end{equation}
For each step, progress reflection provides intermediate signals beyond sparse outcome reward:
\begin{equation}
r_t = r_{\text{env}}(t) + r_p(t) + r_{\text{align}}(t) + \mathbbm{1}[t=T]
\left(
R_{\text{corr}}(\tau) + R_{\text{retro-format}}(\tau)
\right).
\end{equation}
$r_{\text{env}}(t)$ is the environment reward, which is sparse, outcomet-based.
$r_p(t)$ rewards retrospective progress differences, and
$r_{\text{align}}(t)$ encourages pointwise online-retrospective
agreement. The trajectory-level correlation and retrospective
format rewards are applied at the final step.

\begin{equation}
\setlength{\abovedisplayskip}{5pt}
\setlength{\belowdisplayskip}{5pt}
\begin{aligned}
r_p(t) &=
\begin{cases}
\beta (p_{t+1} - p_t), & t < T,\\
0, & t = T,
\end{cases}\\
r_{\text{align}}(t) &=
\frac{\alpha}{T}
\max\left(0, 1 - |p_t - \tilde{p}_t|\right),\\
R_{\text{corr}}(\tau)
&= \alpha_{\text{corr}} \,
\operatorname{PearsonCorr}
\left(
(\tilde{p}_t)_{t=1}^{T},
(p_t)_{t=1}^{T}
\right).
\end{aligned}
\end{equation}
Additionally, the retrospective format reward checks for a valid progress
estimate. The boundary reward encourages the initial online estimate to be $0$ and the last of successful trajectories to be $1$:

\begin{equation}
\setlength{\abovedisplayskip}{5pt}
\setlength{\belowdisplayskip}{5pt}
\begin{aligned}
R_{\text{retro-format}}(\tau)
&=
\begin{cases}
+0.1, & \text{if the retrospective format is valid},\\
-0.1, & \text{otherwise},
\end{cases}\\
R_{\text{boundary}}(\tau)
&= -w_p \left(
|\tilde{p}_1|
+ \mathbbm{1}[\text{success} \land T>1]
|\tilde{p}_T - 1|
\right).
\end{aligned}
\end{equation}

Following GiGPO \citep{feng2026groupingroup}, we compute hierarchical advantages at two granularities.
The episode advantage $A^{\text{episode}}(\tau)$ normalizes the total return $R(\tau)$ against trajectories rolled out from the same task.
The step advantage $A^{\text{step}}_t$ compares discounted returns $G_t$ among actions taken from the same anchor state, i.e., identical observations $o_k = o_t$ that naturally recur across trajectories,
\begin{equation}
\label{l2}
\setlength{\abovedisplayskip}{5pt}
\setlength{\belowdisplayskip}{5pt}
G_t = \sum_{k=0}^{T-t} \gamma^k r_{t+k} = r_t + \gamma G_{t+1}.
\end{equation}

Since $r_t$ incorporates progress shaping, $G_t$ encodes \textit{when} progress increases occur, enabling fine-grained credit assignment.
The final advantage combines both levels and is shared by all tokens
in the forward response, including reasoning, progress estimates,
and actions. We optimize a token-level clipped objective, with
$\rho_{t,\ell}(\theta)
= \frac{\pi_\theta(y_{t,\ell} \mid h_t, y_{t,<\ell})}
{\pi_{\text{old}}(y_{t,\ell} \mid h_t, y_{t,<\ell})}$,
where $y_{t,\ell}$ denotes the $\ell$-th token of the full response
at step $t$:
\begin{equation}
\begin{aligned}
\setlength{\abovedisplayskip}{5pt}
\setlength{\belowdisplayskip}{5pt}
A_t = \underbrace{\frac{R(\tau) - \text{mean}\bigl(\{R(\tau_j)\}_{j=1}^N\bigr)}{F_{\text{norm}}\bigl(\{R(\tau_j)\}_{j=1}^N\bigr)}}_{A^{\text{episode}}(\tau)}
    + \omega \cdot
      \underbrace{\frac{G_t - \text{mean}\bigl(\{G_k \mid o_k = o_t\}\bigr)}{F_{\text{norm}}\bigl(\{G_k \mid o_k = o_t\}\bigr)}}_{A^{\text{step}}_t}
\\
\mathcal{L}_{\text{PG}}(\theta)
= -\mathbb{E}_{\tau,t,l}
\left[
\min\left(
\rho_{t,\ell}(\theta) A_t,\,
\operatorname{clip}\left(
\rho_{t,\ell}(\theta),1-\epsilon,1+\epsilon
\right) A_t
\right)
\right].
\end{aligned}
\end{equation}

\section{Experiments}


\subsection{Setup}
\paragraph{Benchmarks}
We evaluate on three agentic benchmarks:
\textbf{WebShop} for online shopping~\citep{yao2022webshop},
\textbf{ALFWorld} for text-based household tasks~\citep{shridhar2020alfworld},
and \textbf{Sokoban} for box-pushing puzzles~\citep{SchraderSokoban2018}.
We report task success rate (SR) and environment score on 256 test
samples per benchmark. 

\begin{table*}[t]
\vspace{-8pt}
\centering
\small
\setlength{\belowcaptionskip}{3pt}
\caption{Main results on WebShop across three model sizes. 
For training methods, we report both the best score (\textit{Best}) and 
convergence window average (\textit{Avg}). 
SR: success rate (\%). 
$\Delta$: improvement over the corresponding baseline.
The same applies below.
}
\resizebox{\textwidth}{!}{
\begin{tabular}
{l l llll llll llll}
\toprule
& & \multicolumn{4}{c}{\textbf{Qwen2.5-1.5B}} 
& \multicolumn{4}{c}{\textbf{Qwen2.5-3B}} 
& \multicolumn{4}{c}{\textbf{Qwen2.5-7B}} \\
\cmidrule(lr){3-6} \cmidrule(lr){7-10} \cmidrule(lr){11-14}
\textbf{Method} & \textbf{} 
& \textbf{SR} & \textbf{$\Delta$} & \textbf{Score} & \textbf{$\Delta$} 
& \textbf{SR} & \textbf{$\Delta$} & \textbf{Score} & \textbf{$\Delta$}
& \textbf{SR} & \textbf{$\Delta$} & \textbf{Score} & \textbf{$\Delta$}\\
\midrule

\multicolumn{14}{l}{\textit{\textbf{Prompting Methods}}} \\
\rowcolor{weboursblue}
\multicolumn{2}{l}{Base Prompt}  
& 0.4 & -- & 5.0 & -- 
& 12.5 & -- & 36.5 & -- 
& 14.1 & -- & 50.8 & -- \\
\rowcolor{white}
\multicolumn{2}{l}{Meta Prompt for Progress }  
& 1.2 & +0.8 & 3.8 & -1.2 
& 16.4 & +3.9 & 41.5 & +5.0 
& 10.9 & -3.2 & 41.4 & -9.4 \\
\multicolumn{2}{l}{Random Progress}   
& 0.8 & +0.4 & 13.2 & +8.2 
& 16.4 & +3.9 & 52.4 & +15.9 
& 10.2 & -3.9 & 44.4 & -6.4 \\
\multicolumn{2}{l}{Retro Progress}   
& 5.1 & +4.7 & 20.7 & +15.7 
& 10.6 & -1.9 & 44.5 & +8.0 
& 23.8 & +9.7 & 51.5 & +0.7 \\
\multicolumn{2}{l}{Random Retro Progress}   
& 2.0 & +1.6 & 18.5 & +13.5 
& 16.0 & +3.5 & 48.4 & +11.9 
& 12.5 & -1.6 & 46.5 & -4.3 \\

\noalign{\vskip 2pt}
\hdashline
\noalign{\vskip 2pt}
\multicolumn{14}{l}{\textit{\textbf{Training Methods}}} \\

\multirow{2}{*}{GRPO} 
& Best 
& 66.41 & -7.03 & 81.22 & -8.60 
& 50.00 & -21.88 & 71.02 & -13.39
& 73.29 & -5.62 & 82.18 & -3.75 \\
& Avg  
& 62.77 & -4.38 & 77.26 & -8.01 
& 44.53 & -23.63 & 63.90 & -20.48
& 64.02 & -10.63 & 76.46 & -8.63 \\

\multirow{2}{*}{GiGPO} 
& Best 
& 67.97 & -5.47 & 81.35 & -8.47 
& 71.88 & +0.00 & 83.94 & -0.47
& 74.22 & -4.69 & 83.98 & -1.95 \\
& Avg  
& 59.84 & -7.31 & 77.68 & -7.59 
& 63.57 & -4.59 & 81.49 & -2.89
& 71.76 & -2.89 & 83.77 & -1.32 \\

\rowcolor{weboursblue}
& Best 
& 73.44 & -- & 89.82 & -- 
& 71.88 & -- & 84.41 & -- 
& 78.91 & -- & 85.93 & -- \\
\rowcolor{weboursblue}
\multirow{-2}{*}{Meta Prompt} 
& Avg  
& 67.15 & -- & 85.27 & -- 
& 68.16 & -- & 84.38 & -- 
& 74.65 & -- & 85.09 & -- \\

\multirow{2}{*}{L1} 
& Best 
& 68.75 & -4.69 & 84.51 & -5.31 
& 76.56 & +4.68 & 88.53 & +4.12 
& 80.86 & +1.95 & \textbf{91.74} & +5.81 \\
& Avg  
& 63.20 & -3.95 & 82.17 & -3.10 
& 71.68 & +3.52 & 86.21 & +1.83 
& 75.62 & +0.97 & 87.81 & +2.72 \\

\multirow{2}{*}{L2} 
& Best 
& 75.00 & +1.56 & 88.01 & -1.81 
& 76.56 & +4.68 & 88.60 & +4.19 
& 75.39 & -3.52 & 87.54 & +1.61 \\
& Avg  
& 70.47 & +3.32 & 85.67 & +0.40 
& 70.86 & +2.70 & 86.49 & +2.11 
& 70.59 & -4.06 & 83.45 & -1.64 \\

\noalign{\vskip 2pt}
\hdashline
\noalign{\vskip 2pt}
\multicolumn{14}{l}{\textit{\textbf{RePro (Ours)}}} \\

\rowcolor{green}
& Best 
& \textbf{81.64} & +8.20 & \textbf{91.72} & +1.90 
& \textbf{83.59} & +11.71 & \textbf{92.09} & +7.68 
& \textbf{84.38} & +5.47 & 91.18 & +5.25 \\
\rowcolor{green}
\multirow{-2}{*}{RePro} 
& Avg  
& \textbf{76.13} & +8.98 & \textbf{87.56} & +2.29 
& \textbf{79.73} & +11.57 & \textbf{90.75} & +6.37 
& \textbf{80.47} & +5.82 & \textbf{89.13} & +4.04 \\

\bottomrule
\end{tabular}
}
\vspace{-5pt}
\label{tab:mainres-webshop}

\end{table*}

\begin{table*}[t]
\vspace{-3pt}
\centering
\small
\setlength{\belowcaptionskip}{3pt}
\caption{Main results on ALFWorld and Sokoban across three model sizes. 
}
\resizebox{\textwidth}{!}{
\begin{tabular}
{l l llll llll llll}
\toprule
& & \multicolumn{4}{c}{\textbf{Qwen2.5-1.5B on ALFWorld}} 
& \multicolumn{4}{c}{\textbf{Qwen2.5-7B on ALFWorld}} 
& \multicolumn{4}{c}{\textbf{Qwen3-4B on Sokoban}} \\
\cmidrule(lr){3-6} \cmidrule(lr){7-10} \cmidrule(lr){11-14}
\textbf{Method} & \textbf{} 
& \textbf{SR} & \textbf{$\Delta$} & \textbf{Score} & \textbf{$\Delta$} 
& \textbf{SR} & \textbf{$\Delta$} & \textbf{Score} & \textbf{$\Delta$}
& \textbf{SR} & \textbf{$\Delta$} & \textbf{Score} & \textbf{$\Delta$}\\
\midrule

\multicolumn{14}{l}{\textit{\textbf{Training Methods}}} \\

\multirow{2}{*}{GiGPO} 
& Best & 89.84 & +2.34 & 5.74 & +0.10 & 95.31 & +0.00 & 7.82 & -0.15 & 86.72 & +1.56 & 6.21 & +0.27 \\ 
& Avg  & 84.38 & +5.90 & 5.49 & +1.02 & 90.08 & -1.29 & 6.34 & -0.15 & 76.09 & +2.03 & 4.21 & +0.29 \\

\rowcolor{weboursblue} 
& Best & 87.50 & -- & 5.64 & -- & 95.31 & -- & 7.97 & -- & 85.16 & -- & 5.94 & -- \\
\rowcolor{weboursblue} 
\multirow{-2}{*}{Meta Prompt}
& Avg  & 78.48 & -- & 4.47 & -- & 91.37 & -- & 6.49 & -- & 74.06 & -- & 3.91 & -- \\

\multirow{2}{*}{L1} 
& Best & 95.70 & +8.20 & 7.88 & +2.24 & 99.61 & +4.30 & 9.78 & +1.81 & \textbf{89.84} & +4.68 & 6.53 & +0.60 \\
& Avg  & 88.05 & +9.57 & 5.81 & +1.34 & \textbf{97.81} & +6.44 & 8.83 & +2.34 & 79.22 & +5.16 & 4.56 & +0.64 \\

\multirow{2}{*}{L2} 
& Best & 74.22 & -13.28 & 4.43 & -1.21 & 98.83 & +3.52 & 9.29 & +1.32 & 88.28 & +3.12 & 6.44 & +0.50 \\
& Avg  & 65.62 & -12.86 & 3.63 & -0.84 & 96.60 & +5.23 & 8.30 & +1.81 & 80.08 & +6.02 & 4.79 & +0.88 \\

\noalign{\vskip 2pt}
\hdashline
\noalign{\vskip 2pt}
\multicolumn{14}{l}{\textit{\textbf{RePro (Ours)}}} \\

\rowcolor{green}
& Best & \textbf{99.22} & +11.72 & \textbf{9.42} & +3.78 & \textbf{100.00} & +4.69 & \textbf{10.26} & +2.29 & 88.28 & +3.12 & \textbf{6.61} & +0.67 \\
\rowcolor{green}
\multirow{-2}{*}{RePro} 
& Avg  & \textbf{96.02} & +17.54 & \textbf{7.94} & +3.47 & 97.54 & +6.17 & \textbf{8.90} & +2.41 & \textbf{81.64} & +7.58 & \textbf{5.10} & +1.18 \\

\bottomrule
\end{tabular}
}
\vspace{-5pt}
\label{tab:mainres2}
\end{table*}

\begin{table*}[t]
\vspace{-8pt}
\centering
\small
\setlength{\belowcaptionskip}{3pt}
\caption{Results for ablation study and variants of our method, using the Best numbers.}
\resizebox{\textwidth}{!}{
\begin{tabular}{l llll llll llll}
\toprule
& \multicolumn{4}{c}{\textbf{Qwen2.5-1.5B}}
& \multicolumn{4}{c}{\textbf{Qwen2.5-3B}}
& \multicolumn{4}{c}{\textbf{Qwen2.5-7B}} \\
\cmidrule(lr){2-5} \cmidrule(lr){6-9} \cmidrule(lr){10-13}
\textbf{Method}
& \textbf{SR} & \textbf{$\Delta$} & \textbf{Score} & \textbf{$\Delta$}
& \textbf{SR} & \textbf{$\Delta$} & \textbf{Score} & \textbf{$\Delta$}
& \textbf{SR} & \textbf{$\Delta$} & \textbf{Score} & \textbf{$\Delta$} \\
\midrule
\multicolumn{13}{l}{\textit{\textbf{RePro (Ours)}}} \\
\rowcolor{weboursblue}
RePro
& 81.64 & -- & 91.72 & --
& 83.59 & -- & 92.09 & --
& 84.38 & -- & 91.18 & -- \\

\noalign{\vskip 2pt}
\hdashline
\noalign{\vskip 2pt}
\multicolumn{13}{l}{\textit{\textbf{Ablations}}} \\

w/ 7B Warmup data
& 83.59 & +1.95 & 93.14 & +1.42
& 82.81 & -0.78 & 91.28 & -0.81
&-- & -- & -- & -- \\

Warmup + GiGPO
& 75.00 & -6.64 & 88.44 & -3.28
& 78.91 & -4.68 & 87.94 & -4.15
& 79.69 & -4.69 & 88.38 & -2.80 \\

Warmup Only
& 0.00 & -81.64 & 0.00 & -91.72
& 28.52 & -55.07 & 32.49 & -59.60
& 61.72 & -22.66 & 71.94 & -19.24 \\

w/o Warmup
& NA & NA & NA & NA
& NA & NA & NA & NA
& 80.08 & -4.30 & 89.96 & -1.22 \\

\noalign{\vskip 2pt}
\hdashline
\noalign{\vskip 2pt}
\multicolumn{13}{l}{\textit{\textbf{Variants}}} \\

L1+ Reshaping
& 72.27 & -9.37 & 87.43 & -4.29
& 72.66 & -10.93 & 84.82 & -7.27
& 80.08 & -4.30 & 87.54 & -3.64 \\

L1+ Reweighting
& 71.48 & -10.16 & 83.55 & -8.17
& 75.00 & -8.59 & 87.62 & -4.47
& 77.34 & -7.04 & 88.98 & -2.20 \\

L1+ Grouping
& 75.78 & -5.86 & 89.63 & -2.09
& 73.83 & -9.76 & 86.32 & -5.77
& 78.13 & -6.25 & 89.23 & -1.95 \\

\bottomrule
\end{tabular}
}
\label{tab:ws—ablation-var}
\end{table*}

\paragraph{Baselines}
Baseline methods for comparison are twofold.
\textbf{(1) Prompting methods.}
Following our pilot study, we evaluate online and retrospective progress prompting, using random progress as a control for the added prompt structure.
\textbf{(2) Training methods.}
We use standard GRPO~\cite{deepseekmath} and GiGPO~\cite{zheng2025group} as our primary RL training baselines.
We further include a progress-aware prompting baseline, denoted as \textit{Meta Prompt}, where the agent is explicitly prompted to assess online progress.
Since \textit{Meta Prompt} uses the same progress-aware prompting format as our method, we use it as the reference baseline for computing $\Delta$ across methods.
On the basis of \textit{Meta Prompt}, two vanilla implementations of progress augmentation are evaluated.
\textit{L1} denotes only adding a format penalty in the reward function to the \textit{Meta Prompt}, while \textit{L2} denotes directly using $\hat{p}_t$ as the step reward $r_t$ in Equation \ref{l2}.
Full details are provided in Appendix~\ref{app:implementation}.

\subsection{Main Results}

\paragraph{Progress assessment does not reliably emerge from prompting or outcome-oriented training.}
As shown in Table~\ref{tab:mainres-webshop}, neither improving prompting nor existing training methods reliably elicits effective progress assessment.
Simple prompting often degrades performance, while retrospective prompting without training yields only limited gains.
Regarding training methods, \textit{Meta Prompt} induces longer reasoning traces and yields performance gains in most settings compared with GRPO and GiGPO, indicating that progress-aware prompting provides a richer context, with additional tokens not necessarily becoming a burden.
We further observe that outcome-oriented RL with progress prompt and simple leverage (L1/L2) produces inconsistent improvements across settings. 

\paragraph{RePro improves task success across model sizes and environments.}
Table~\ref{tab:mainres-webshop} shows that RePro consistently outperforms all baselines on WebShop across 1.5B, 3B, and 7B models, achieving absolute success rate gains of +8.98\%, +11.57\%, and +5.82\% over the \textit{Meta Prompt} baseline, respectively.
Compared to progress-related variants, RePro further improves performance, demonstrating the effectiveness of retrospective progress modeling beyond vanilla reward engineering.
Across environments, RePro generalizes to both ALFWorld household planning (+11.72 SR for Qwen2.5-1.5B and +4.69 SR for Qwen2.5-7B) and Sokoban spatial reasoning (+3.12 SR for Qwen3-4B), supporting the hypothesis that retrospective progress awareness enhance long-horizon agent learning.
The improvement on Sokoban is relatively smaller, which may be because its reward already provides partial process-oriented feedback.

\begin{figure*}[t]
    \centering
    \includegraphics[width=0.95\textwidth]{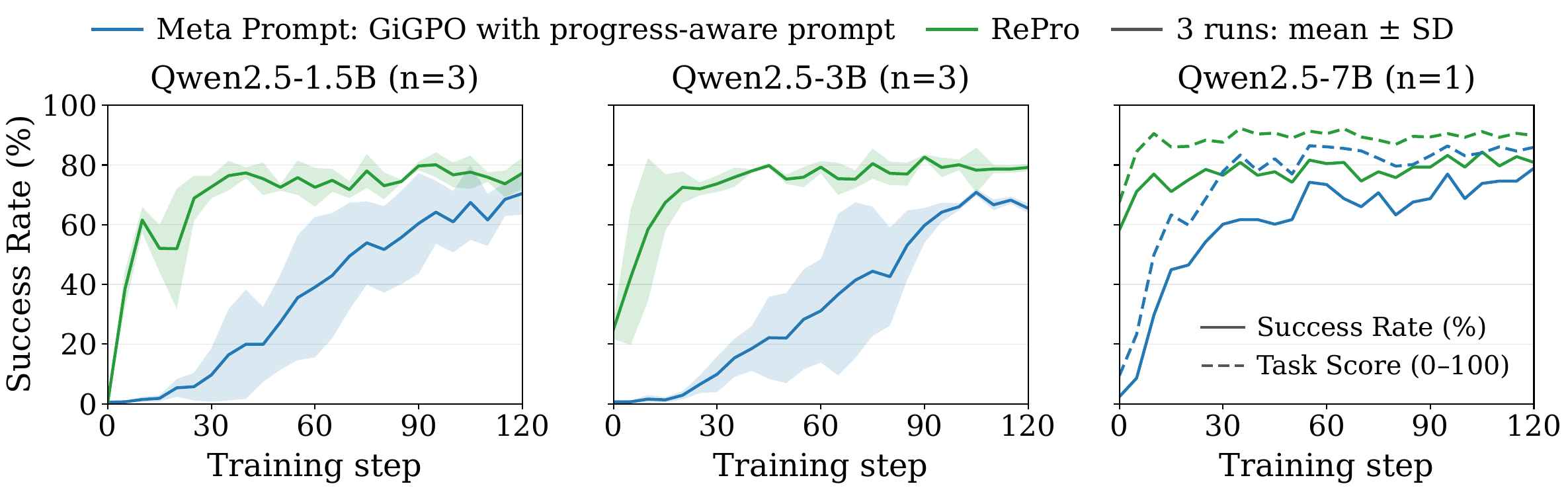}
    \vspace{-5pt}
    \caption{Learning curves on WebShop. Solid lines show evaluation success rate; shaded regions indicate standard deviation across three runs. The dashed curves in the 7B panel show task score.}
    \label{fig:curve}
    \vspace{-5pt}
\end{figure*}

\paragraph{Across-seed consistency.}
To assess sensitivity to training randomness, we repeat WebShop training with three independent seeds for Qwen2.5-1.5B and Qwen2.5-3B. Figure~\ref{fig:curve} and Table~\ref{tab:critic-comparison} report the mean validation performance with standard deviation across runs. 
RePro reaches a high success rate earlier than the \textit{Meta Prompt} baseline and maintains an advantage throughout training. After step 100, the success-rate gap remains approximately 10 percentage points on average across the plotted checkpoints for both model sizes. The task-score curves show the same qualitative pattern. For Qwen2.5-7B, we report one training run with both success rate and task score, showing a similar trend of faster convergence and an advantage during the later training stages. 

\section{Analysis}
\subsection{Ablations and Variants}
Ablation settings for RePro include:
(1) w/ 7B warmup data: Using the warmup dataset from 7B trajectories.
(2) Warmup + GiGPO: Two-phase training with SFT warmup followed by standard GiGPO without retrospective progress mechanism.
(3) Warmup Only: Performance after only warmup on retrospective-labeled data without subsequent RL training. 
(4) w/o Warmup: Direct RL from base model with retrospective mechanism enabled.
The following are variants that leverage progress assessments in different ways.
Table~\ref{tab:ws—ablation-var} reports the ablations and variants.

\textbf{(1) The Role of SFT Warmup.}
\textit{Warmup Only} achieves substantially lower performance than RePro (crashing on 1.5B and 3B, -22.66\% on 7B), despite achieving nearly perfect format accuracy, showing that SFT teaches the \emph{format} of retrospective reflection but does not yield effective progress awareness without task-reward grounding.
Removing warmup destabilizes training for 1.5B and 3B models, indicating that warmup is critical for establishing the forward-then-reflect behavior. Using stronger warmup demonstrations from the 7B model further improves the 1.5B model.
\textbf{(2) The Role of Retrospective Reflection.}
Comparisons against \textit{Warmup + GiGPO} and \textit{L1 + Reshaping} show that retrospective reflection contributes beyond format initialization or reward shaping alone.
Retrospective progress differences correlate positively with step-level advantages in 91.5\% of training rollouts, an association that partly reflects the use of progress shaping in advantage computation.

\begin{table}[t]
\vspace{-8pt}
\centering
\small
\setlength{\belowcaptionskip}{3pt}
\setlength{\tabcolsep}{5pt}
\caption{
Progress assessment analysis on WebShop.
IntermDisc is reported in percentage points, while temporal $R^2$ is scaled
by 100, and all other metrics are percentages (Appendix~\ref{app:progress_metrics}).
}
\begin{tabular}{@{}ll rr rr rr}

\toprule
& & \multicolumn{2}{c}{\textbf{Qwen2.5-1.5B}}
  & \multicolumn{2}{c}{\textbf{Qwen2.5-3B}}
  & \multicolumn{2}{c}{\textbf{Qwen2.5-7B}} \\
\cmidrule(lr){3-4}
\cmidrule(lr){5-6}
\cmidrule(lr){7-8}
\textbf{Metric} & \textbf{Outcome}
& Baseline & \cellcolor{weboursblue}\textbf{RePro}
& Baseline & \cellcolor{weboursblue}\textbf{RePro}
& Baseline & \cellcolor{weboursblue}\textbf{RePro} \\
\midrule

\textbf{IntermDisc (pp)} & --
& 1.86  & \cellcolor{weboursblue}\textbf{6.37}
& 1.11  & \cellcolor{weboursblue}\textbf{31.62}
& $-0.01$ & \cellcolor{weboursblue}\textbf{19.34} \\
\midrule

\multirow{2}{*}{Temporal $R^2$}
& Success
& 96.89 & \cellcolor{weboursblue}78.14
& 21.62 & \cellcolor{weboursblue}94.71
& 48.74 & \cellcolor{weboursblue}96.52 \\
& Failure
& 97.02 & \cellcolor{weboursblue}75.64
& 18.69 & \cellcolor{weboursblue}60.72
& 46.13 & \cellcolor{weboursblue}79.24 \\
\addlinespace[3pt]

\multirow{2}{*}{Monotonicity (\%)}
& Success
& 99.99 & \cellcolor{weboursblue}99.60
& 99.34 & \cellcolor{weboursblue}100.00
& 99.98 & \cellcolor{weboursblue}99.99 \\
& Failure
& 100.00 & \cellcolor{weboursblue}99.26
& 99.32  & \cellcolor{weboursblue}92.76
& 99.94  & \cellcolor{weboursblue}95.72 \\
\addlinespace[3pt]

\multirow{2}{*}{Decline (\%)}
& Success
& 0.04 & \cellcolor{weboursblue}1.28
& 2.72 & \cellcolor{weboursblue}0.03
& 0.15 & \cellcolor{weboursblue}0.02 \\
& Failure
& 0.00 & \cellcolor{weboursblue}2.88
& 3.01 & \cellcolor{weboursblue}30.58
& 0.42 & \cellcolor{weboursblue}20.07 \\
\addlinespace[3pt]

\multirow{2}{*}{Plateau (\%)}
& Success
& 0.47  & \cellcolor{weboursblue}27.08
& 78.49 & \cellcolor{weboursblue}1.37
& 99.98 & \cellcolor{weboursblue}0.45 \\
& Failure
& 0.57  & \cellcolor{weboursblue}33.03
& 82.18 & \cellcolor{weboursblue}12.77
& 99.86 & \cellcolor{weboursblue}6.45 \\

\bottomrule
\end{tabular}

\label{tab:progress_combined}
\end{table}

\subsection{Progress Predictions}

\paragraph{Metrics.}
Progress lacks objective ground truth in outcome-based agentic tasks, making direct accuracy evaluation infeasible.
We therefore examine outcome alignment and behavioral properties.
Our primary metric, \textbf{Intermediate Discrimination (IntermDisc)},
measures the average stepwise gap in progress estimates between
successful and failed trajectories.
We additionally report \textbf{Temporal $R^2$},
\textbf{Monotonicity}, \textbf{Decline Rate}, and
\textbf{Plateau Rate} to characterize progress dynamics.
Detailed definitions and aggregation procedures are provided in
Appendix~\ref{app:progress_metrics}.

\paragraph{Findings.}
Table~\ref{tab:progress_combined} compares the baseline's progress estimates with RePro's retrospective estimates, which serve as reward signals during training. Both are evaluated on their respective test-set trajectories.
\noindent\textbf{(1) RePro's progress signals distinguish task outcomes more clearly.}
Baseline IntermDisc remains close to zero across model sizes,
ranging from -0.01 to 1.86 percentage points.
RePro achieves higher discrimination:
6.37 (1.5B), 31.62 (3B), and 19.34 (7B).
The stepwise profiles in Appendix~\ref{app:progress_metrics},
Figure~\ref{fig:discrim}, show corresponding separation
between successful and failed trajectories, particularly for 3B
and 7B; the gap is smaller for 1.5B.
These results show that RePro's retrospective supervision
differentiates progress according to the eventual task outcome.
\noindent\textbf{(2) Progress dynamics also differ across outcomes.}
With RePro, successful trajectories exhibit higher temporal $R^2$
and monotonicity than failed trajectories across all three model sizes.
Decline rates on failed trajectories are increased (e.g., 30.58\% to 0.03\%) while plateau rates decrease (e.g., 82.18\% to 12.77\%).
RePro's retrospective signals exhibit more
downward revisions on failed trajectories.
Representative progress assessments appear in
Table~\ref{tab:qualitative_progress}
(Appendix~\ref{app:qualitative_cases}).

\section{Conclusion}
This work investigates how LLM-based agents can turn interaction histories into training signals that complement sparse outcome rewards.
We introduce RePro, which uses the acting policy's own retrospective reasoning over completed trajectories and known outcomes to generate step-wise progress assessments.
Following supervised warmup, these assessments provide progress-based reward shaping and alignment targets for online estimates, without requiring a separate process reward model or ongoing external annotation.
Experiments on WebShop, ALFWorld, and Sokoban show that this approach improves task success across multiple model settings.
These findings support retrospective progress assessment as auxiliary supervision for long-horizon tasks and motivate further research into self-improving agents.


\section*{AI Use Statement}
LLMs were used solely in an auxiliary capacity, primarily for linguistic refinement, including grammar correction, clarity improvement, phrasing and formulation checks, numerical consistency checks, and the removal of non-academic expressions. 
We also used DeepSeek to generate retrospective progress labels for the supervised warmup data.
LLMs were \textbf{not} used to generate research ideas, design experiments, conduct literature reviews, or make methodological decisions. All conceptual contributions, experimental designs, methodological choices, and research conclusions were independently developed and carried out by the authors. We have reviewed all AI-assisted work.

\section*{Ethics Statement}
This work studies LLM-based agent training in controlled benchmark environments, including WebShop, ALFWorld, and Sokoban.
The experiments do not involve human evaluation, real-world purchases, or physical actions.

\section*{Reproducibility Statement}
Sections~3-5 describe the task formulation, training method,
and evaluation setup, with results and ablations in Sections~5-6.
Appendix~\ref{app:implementation} provides the retrospective
reflection prompt, the pilot evaluation protocol and demonstration construction, training configurations, reward implementation,
and checkpoint and multi-seed reporting protocols.
Appendix~\ref{app:progress_metrics} defines the progress-assessment
metrics and their aggregation procedures.
\bibliographystyle{iclr2027_conference}

\bibliography{custom}

\appendix



\section{Additional Positioning of RePro}
\label{app:positioning}

\noindent\textbf{Relation to actor-critic learning, process supervision, and reward shaping.}
A learned critic and RePro's progress-derived rewards can both facilitate credit assignment, but serve different roles.
A critic trained on environment rewards usually estimates the expected discounted return:
\begin{equation}
V_{\mathrm{env}}^\pi(h_t)
=
\mathbb{E}_{\pi}
\left[
\sum_{k=t}^{T}\gamma^{k-t}r_{\mathrm{env}}(k)
\;\middle|\; h_t
\right].\end{equation}
Although the critic may learn representations correlated with task progress, 
its training targets do not explicitly describe completed subgoals or explain how individual interactions advance task completion.
RePro instead verbalizes such reasoning and augments the reward function using progress assessments generated retrospectively from the complete trajectory and realized outcome:
\begin{equation}
(p_1,\ldots,p_T)
\sim \pi_\theta^{\mathrm{retro}}(\cdot\mid\tau,r).
\end{equation}
These assessments combine hindsight context unavailable at the corresponding online step with the semantic knowledge of LLM itself and verbalized reasoning about task completion to infer implicit task structure.
For example, in WebShop, it can recognize that a search or filtering action completes an information-gathering subgoal even when the environment provides no immediate reward.
Consequently, trajectories with the same terminal outcome may receive different intermediate progress supervision according to how their actions advance task completion.
Online-retrospective alignment further trains the policy's online estimates toward these hindsight-derived targets.

RePro is also related to process supervision, but does not require human-labeled intermediate steps or a separately trained process reward model~\citep{lightman2024lets,zou2026reasonfluxprm,xi2026agentprm}.
Retrospection Warmup uses a small externally generated set of successful trajectories to initialize the reflection format; subsequent progress assessments in RePro-PO are generated by the agent itself from completed trajectories and task outcomes.
RePro further builds on reward shaping.
Unlike shaping signals constructed from manually specified subgoals or environment-specific state changes, RePro derives its signal from differences between language-based retrospective progress assessments and combines it with online-retrospective alignment.


Empirically, the GiGPO study provides context for our choice of a critic-free optimizer: it reports stronger WebShop performance than PPO and GRPO under its experimental setup~\citep{feng2026groupingroup}.
Our experiments evaluate the additional benefit of RePro within a GiGPO-based training setup.
Table~\ref{tab:critic-comparison} reports published results alongside our comparison between RePro and the GiGPO-based Meta Prompt baseline.

\begin{table*}[ht]
\centering
\small
\setlength{\belowcaptionskip}{3pt}
\caption{Contextual comparison of critic-based, critic-free, and progress-aware optimization on WebShop. Results are from the GiGPO study and our experiments.
For our experiments, we report the mean and standard deviation
across three seeds, using each run's average success rate over
its final five evaluation checkpoints.}
\begin{tabular}{llll}
\toprule
Model & Method & WebShop success rate & source \\
\midrule
\multirow{5}{*}{Qwen2.5-1.5B}
    & PPO   & $51.5 \pm 2.9$ & \cite{feng2026groupingroup} \\
    & GRPO  & $56.8 \pm 3.8$ & \cite{feng2026groupingroup}  \\
    & GiGPO & $67.4 \pm 4.5$ & \cite{feng2026groupingroup}   \\
    & GiGPO + Meta Prompt & $65.78 \pm 8.68$ & Figure \ref{fig:curve} \\
    & RePro & $76.22 \pm 1.20$ & Figure \ref{fig:curve} \\
\bottomrule
\end{tabular}
\label{tab:critic-comparison}
\end{table*}

These results do not establish that retrospective progress supervision is fundamentally or information-theoretically superior to a learned critic.
Rather, the published results provide context for critic-based and
critic-free optimization under a comparable experimental setup and our experiments show that explicit progress supervision provides an effective additional signal within the critic-free GRPO/GiGPO framework.
A critic could in principle be trained using RePro's progress-aware rewards: this would combine the two approaches rather than make retrospective progress supervision redundant.
The choice between critic-based and critic-free optimization is therefore orthogonal to RePro's main contribution.

\noindent\textbf{Relation to reflection and self-evolution.}
Inference-time reflection methods convert past trajectories into verbal feedback, episodic memory, reusable heuristics, or refined plans without updating model parameters~\citep{shinn2023reflexion,kim2025reflact,allard2026experiential,wang2026preflect}.
RePro instead uses retrospective reflection as training supervision: the agent first executes the task using online progress estimates and then reassesses the completed trajectory with knowledge of its outcome, allowing the online progress-assessment capability to be internalized through parameter updates.
RePro is also related to training-time self-improvement and self-evolution methods~\citep{qu2024recursive,li2025reflectevo,bensal2025reflect,kumartraining,zhang2026consistent,ou2026serl}.
While these methods broadly optimize self-correction, consistency, or reusable experience, RePro specifically trains outcome-grounded task-progress assessment and integrates the resulting signal into policy optimization.

\section{Implementation Details}
\label{app:implementation}
\subsection{Prompt}
\begin{prompt}[title=Retrospective Progress Reflection Prompt]
\label{prompt:retro}
\small
\texttt{You have just completed a task on an online shopping website. 
Now review the trajectory with the benefit of hindsight.\\
\\
Your trajectory (4 steps):\\
\quad Step 1: action="search[wireless mouse under 20 dollars]" \quad your\_estimate=10\%\\
\quad Step 2: action="click[item - Logitech wireless mouse]" \quad your\_estimate=35\%\\
\quad Step 3: action="click[add to cart]" \quad your\_estimate=70\%\\
\quad Step 4: action="click[buy now]" \quad your\_estimate=90\%\\
\\
Final outcome: SUCCEEDED (score=1.00)\\
\\
Re-assess what fraction of the task was genuinely completed at each step.\\
\\
Guidelines:\\
- Step 1 should be near 0 (task just started).\\
- If SUCCEEDED, the last step should be near 100.\\
- Values should generally be non-decreasing.\\
\\
First reason briefly in <think>...</think>, then output exactly 4 integers (0--100) in 
<hindsight\_progress>v1,v2,v3,v4 \\</hindsight\_progress>.\\
Example: <think>step 1 started, step 2 found item, step 3 bought it.</think>\\<hindsight\_progress>0,60,100\\</hindsight\_progress>}
\end{prompt}

\section{Pilot Study Details}
\label{app:pilot_details}

\paragraph{Evaluation setup.}
We evaluate DeepSeek V4.1-Flash through the official DeepSeek API with thinking disabled, and GPT-5.1 through OpenRouter with Azure as the sole provider.
All evaluations use temperature $0$ and a maximum of 30 environment actions per task.
We fix 256 tasks sampled without replacement from the 500-task WebShop test split using seed 42.

\paragraph{Demonstration construction.}
For each model, we collect 100 trajectories from the separate training split and retain successful ones for annotation.
The same model receives each complete successful trajectory and its terminal reward, and annotates the observation following each action with a percentage and a short rationale.
The resulting demonstration pools contain 30 trajectories for DeepSeek and 32 for GPT-5.1.
For each test-time call, we sample 2 trajectories from the corresponding model's pool.
Each demonstration contains its instruction and chronologically ordered observation-progress pairs; action fields stored in the trajectory files are not explicitly included in the demonstration prompt to avoid extra hints.

\paragraph{Inference protocol.}
The baseline predicts actions without an explicit progress-assessment call.
Direct assessment receives the current observation and accumulated interaction messages, including previous assessments.
Retrospectively guided assessment instead receives the task instruction, current observation, and sampled demonstrations.
Both produce a percentage and a short rationale, which are appended to the current observation in a separate action-prediction call.
This action call retains the interaction history and previous assessments but does not receive demonstrations.

\paragraph{Random controls.}
For \textit{Random Progress}, we independently sample a percentage from
$\{0,5,\ldots,100\}$ at each step and ask the model to generate a task-specific rationale supporting that assigned value.
For \textit{Random Demonstration}, we replace demonstration annotations with pre-generated random percentages and fabricated rationales while preserving the instructions, observations, and their temporal order.

\paragraph{Repeated runs and reporting.}
Figure~\ref{fig:Pilot} summarizes runs with 2 seeds.
The test tasks and demonstration pools are fixed;  the seed initializes demonstration sampling and online random percentages.
API and infrastructure failures are retried or resumed rather than counted as task failures.
In the qualitative example, the two methods share an action-observation prefix, but their earlier progress messages differ, illustrating a behavioral difference.

\subsection{Training Hyper-parameters}
(1) SFT Warmup (Phase 1).
We construct hindsight-labeled data as described in \S\ref{sec:retrospective}:
\begin{itemize}[itemsep=0pt]
\item \textbf{Data source}: Successful trajectories (score $\geq 0.9$ for 
WebShop, success=True for ALFWorld) from pre-trained GiGPO checkpoints
\item \textbf{Hindsight labeling}: DeepSeek-v4 API (temperature=0) generates 
hindsight progress sequences
\item \textbf{Training}: Standard SFT with learning rate $5 \times 10^{-6}$, 
batch size 8, 3 epochs.
\end{itemize}

(2) RL Training (Phase 2).
Following GiGPO~\citep{feng2026groupingroup} with modifications for progress reflection:
\begin{itemize}[itemsep=0pt]
\item \textbf{Rollout}: 8 trajectories per task. Episodes are limited to 15 interaction steps for WebShop and Sokoban, and 50 for ALFWorld.
\item \textbf{PPO hyperparameters}: $\epsilon_{\text{clip}} = 0.2$, $\gamma = 0.95$, and KL coefficient $\beta_{\text{KL}} = 0.01$.
\item \textbf{Learning rate}: $1 \times 10^{-6}$.
\item \textbf{Training steps}: 120/150 iterations.


\item \textbf{Progress reward weights}:
We use $\beta=0.1$ for retrospective progress differences,
$\alpha=0.2$ for pointwise online-retrospective alignment,
$\alpha_{\mathrm{corr}}=0.3$ for correlation,
and $w_p=0.2$ for online boundary constraints.
The retrospective format reward is $+0.1$ for a valid format
and $-0.1$ otherwise.

\item \textbf{Advantage estimation}: We use GiGPO with
$F_{\text{norm}}=1$ and step-advantage weight $\omega=1$.
The episode baseline is computed over step samples within each
task group, with the episode reward repeated for each valid step.

\item \textbf{Policy optimization}: All generated tokens in each
forward response share the corresponding step advantage.
The policy loss is averaged over non-padding response tokens,
including reasoning, progress estimates, and actions.
We use dual clipping with $c=3$ for negative advantages.
\end{itemize}
All RL experiments use full-parameter fine-tuning. We train on PPU using the veRL framework \citep{sheng2024hybridflow}.

\noindent\textbf{Reward implementation.}
Progress percentages are normalized to $[0,1]$ before reward
computation.
The correlation reward uses valid online-retrospective pairs
and is zero when fewer than three pairs are available.
If the retrospective sequence has standard deviation below
$0.01$, the reward is also zero.
Otherwise, an online standard deviation below $0.01$ incurs
a penalty of $-0.3$; remaining cases receive
$\alpha_{\mathrm{corr}}$ times the Pearson correlation.

The retrospective format check requires a
\texttt{<hindsight\_progress>} field containing at least one
parseable numeric value.
Extra values are truncated, and short sequences are extended
by repeating the final value.
Parsed percentages are clipped to $[0,100]$ before normalization.
The format check does not require an exact sequence length
or validate the reasoning content.

\subsection{Setup Details}
\noindent\textbf{Checkpoint selection and averaging.}
To reduce sensitivity to fluctuations in individual evaluations,
we report \textit{Avg} over a convergence window comprising ten
consecutive evaluation checkpoints, selected using the rule below.
\textit{Best} reports the checkpoint with the highest
evaluation success rate, together with the task score at that
same checkpoint.
\textit{Avg} reports equal-weight averages of success rate and
task score over a ten-checkpoint reporting window.
Starting from the final ten evaluated checkpoints, we shift the
window backward by one checkpoint whenever its final success
rate is more than 2\% below the mean of the preceding nine
checkpoints. We stop when this condition no longer holds or
the window reaches the beginning of the evaluation sequence.

\noindent\textbf{Multi-seed reporting.}
Figure~\ref{fig:curve} presents the learning curves across seeds
0, 1, and 2 for the 1.5B and 3B models.
For the 1.5B comparison in Table~\ref{tab:critic-comparison},
we first average success rates over the final five evaluation
checkpoints of each run, then report the mean and standard
deviation across the three seeds.
This aggregation is separate from the windows used for the
main tables and the progress-metric analysis.

\noindent\textbf{Model size and model-version effects.}
For every model environment pair, RePro and its baselines use the same backbone and experimental setup.
Across tasks, model scale, and version, RePro demonstrates consistent improvement with quantitative variations.
The stable training of Qwen2.5-7B without Retrospection Warmup provides further support for our design.
It suggests that stronger models can already perform the forward-then-reflect format, whereas warmup primarily bootstraps this behavior for smaller models.
Relative to Meta Prompt, RePro improves average success rate
by 5.82 percentage points on WebShop and 6.17 points on ALFWorld
for Qwen2.5-7B.
On WebShop, retaining warmup further improves best success rate
by 4.30 points over the variant without warmup, indicating that
warmup remains beneficial even when training without it is stable.

We select Qwen2.5 as the principal backbone family because its matched 1.5B, 3B, and 7B checkpoints enable a controlled scaling study~\citep{qwen2025qwen25technicalreport,feng2026groupingroup}.
The Qwen3-4B result on Sokoban further provides cross-version evidence.

\begin{figure*}[t]
    \centering
    \includegraphics[width=0.95\textwidth]{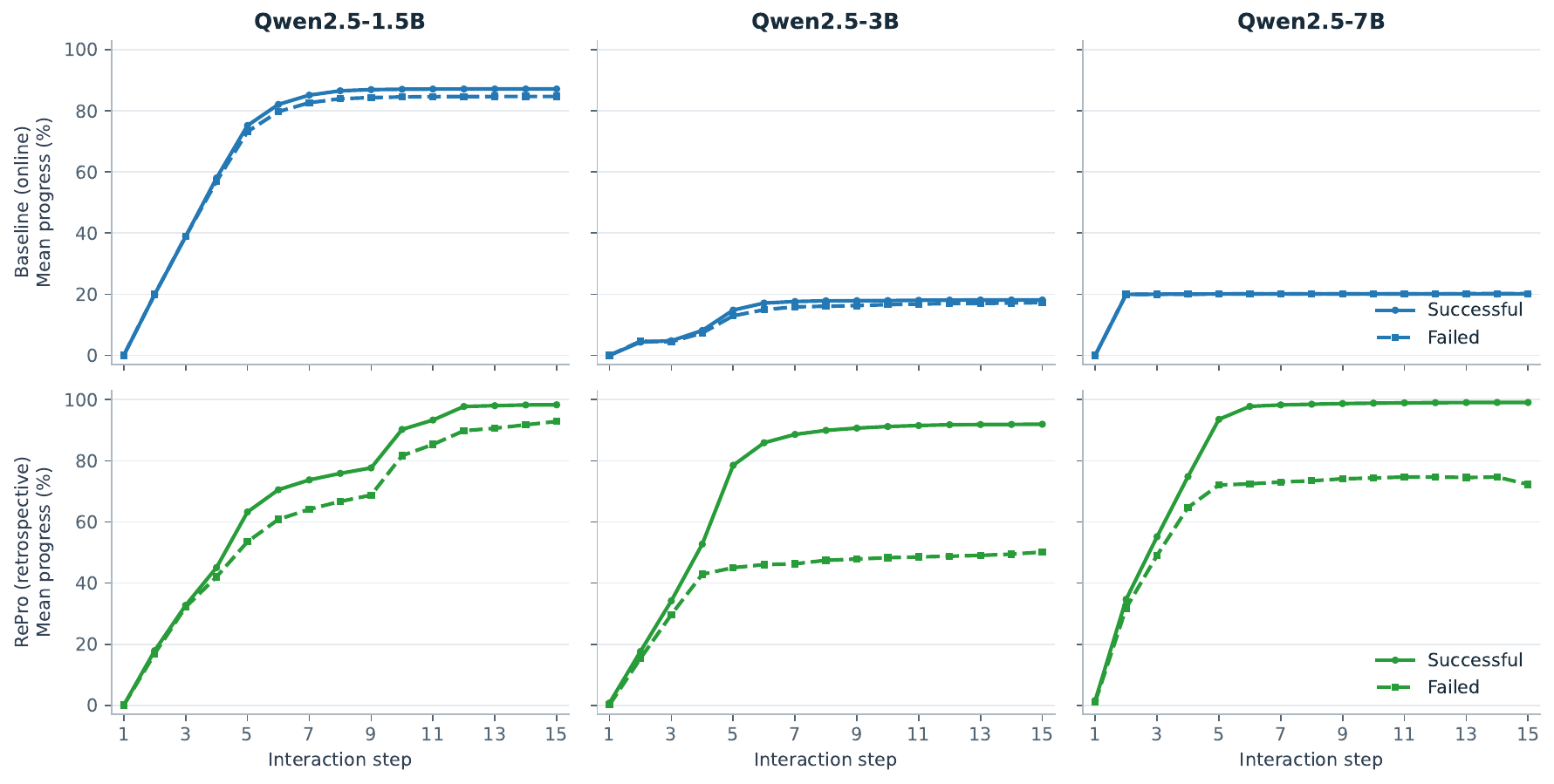}
    \caption{Progress estimates on successful and failed WebShop trajectories.
The top row shows online estimates from the Meta Prompt baseline;
the bottom row shows retrospective estimates from RePro.
Solid and dashed lines denote successful and failed trajectories, respectively.
Short trajectories are padded to 15 steps using their last valid estimate.
Curves are averaged across checkpoints in the same convergence windows
used in Appendix \ref{app:progress_metrics} and Table~\ref{tab:progress_combined}.}
    
    \label{fig:discrim}
\end{figure*}

\section{Progress Metrics}
\label{app:progress_metrics}

\paragraph{Evaluation setup.}
We analyze the baseline's online progress estimates and RePro's
retrospective estimates on their respective test-set evaluation
trajectories.
Let $p_{e,t}\in[0,100]$ denote the progress estimate at step $t$
of trajectory $e$, and let $T_e$ denote its length.
Successful and failed trajectories are grouped by terminal reward
($r_e \geq 9$ and $r_e < 1$, respectively); intermediate-reward
trajectories are excluded from these groups.
The definitions below apply at each evaluation checkpoint.

\paragraph{Intermediate Discrimination (IntermDisc).}
IntermDisc measures the average stepwise progress gap between
successful and failed trajectories.
We retain the first $H=15$ steps of each trajectory and pad shorter
trajectories with their last valid estimate.
Missing estimates within the original sequence remain missing;
trajectories without any valid estimate in the retained sequence
are excluded.
Let $\mu_{\mathrm{s},t}$ and $\mu_{\mathrm{f},t}$ denote the mean
available estimates at step $t$ for successful and failed
trajectories, respectively.
Then
\begin{equation}
\mathrm{IntermDisc}
=
\frac{1}{|\mathcal{T}|}
\sum_{t\in\mathcal{T}}
\left(\mu_{\mathrm{s},t}-\mu_{\mathrm{f},t}\right),
\end{equation}
where $\mathcal{T}\subseteq\{1,\ldots,H\}$ contains steps with
available estimates in both groups.
IntermDisc is reported in percentage points.
The progress profiles use the same stepwise means and padding
procedure.

\paragraph{Temporal $R^2$.}
For each trajectory, we fit a linear regression with an intercept
from relative step position $t/T_e$ to valid progress estimates.
Temporal $R^2$ is the resulting coefficient of determination,
equivalently
\begin{equation}
R_e^2
=
\operatorname{corr}^{2}
\left(t/T_e,\,p_{e,t}\right),
\end{equation}
computed over valid steps.
Trajectories with fewer than three valid estimates or an undefined
$R^2$ are omitted from this metric.
This statistic measures linear association, not its direction
or progress accuracy.

\paragraph{Monotonicity.}
For the remaining metrics, let
$q_{e,1},\ldots,q_{e,n_e}$ be the valid progress estimates in
their original order, and define
$d_{e,j}=q_{e,j+1}-q_{e,j}$.
We use the original sequences without truncation or padding and
exclude trajectories with fewer than two valid estimates.
Monotonicity is the fraction of non-decreasing transitions:
\begin{equation}
\mathrm{Mono}_e
=
\frac{1}{n_e-1}
\sum_{j=1}^{n_e-1}
\mathbf{1}\!\left[d_{e,j}\geq 0\right].
\end{equation}
Consecutive valid estimates are compared even when missing
estimates occur between them.

\paragraph{Decline Rate.}
A trajectory exhibits a decline if at least one transition
decreases by more than 10 percentage points:
\begin{equation}
D_e
=
\mathbf{1}\!\left[
\min_{1\leq j<n_e}d_{e,j}<-10
\right].
\end{equation}
Decline Rate is the fraction of eligible trajectories with $D_e=1$.

\paragraph{Plateau Rate.}
A trajectory exhibits a plateau if at least half of its valid
transitions change by less than one percentage point:
\begin{equation}
P_e
=
\mathbf{1}\!\left[
\frac{1}{n_e-1}
\sum_{j=1}^{n_e-1}
\mathbf{1}\!\left[|d_{e,j}|<1\right]
\geq \frac{1}{2}
\right].
\end{equation}
Plateau Rate is the fraction of eligible trajectories with $P_e=1$.

\paragraph{Aggregation.}
At each checkpoint, temporal $R^2$, monotonicity, decline, and
plateau are averaged over eligible trajectories separately for
successful and failed outcomes.
Undefined values are excluded.
Table~\ref{tab:progress_combined} reports temporal $R^2$ multiplied
by 100 and the three remaining metrics as percentages.

We then average checkpoint-level results within each run's
convergence window, assigning equal weight to each checkpoint.
For a run with $K$ evaluated checkpoints in training order,
the reference success rate is the mean over the final
$\max(2,\lfloor K/3\rfloor)$ checkpoints.
The convergence window begins at the first checkpoint whose
three-checkpoint forward average success rate reaches at least
$80\%$ of this reference, and includes all subsequent checkpoints.
If no such checkpoint exists, we use the reference tail instead.
Unavailable checkpoint-level values are omitted.
The plotted progress profiles use the same checkpoint windows
and equal-weight averaging.

\section{Qualitative Progress-Assessment Cases}
\label{app:qualitative_cases}

To complement the aggregate progress metrics, we present three representative WebShop trajectories generated by RePro in Table \ref{tab:qualitative_progress}, covering successful progress tracking and two distinct failure patterns.
These examples illustrate model behavior rather than quantify the frequency of each pattern.

\begin{table}[t]
\centering
\small
\renewcommand{\arraystretch}{1.15}
\caption{Representative successful and unsuccessful progress assessments on WebShop.}
\setlength{\belowcaptionskip}{3pt}
\begin{tabularx}{\linewidth}{@{}p{2.2cm}Xp{3.4cm}@{}}
\toprule
\textbf{Case}
& \textbf{Trajectory and generated progress}
& \textbf{Interpretation} \\
\midrule

Successful progress tracking
&
Search (0\%)
$\rightarrow$ click product (10\%)
$\rightarrow$ select color blue/gray (33.33\%)
$\rightarrow$ select size 1'9'' $\times$ 7'2'' (50\%)
$\rightarrow$ reconfirm color (50\%)
$\rightarrow$ reconfirm size (50\%)
$\rightarrow$ buy after all constraints are aligned (100\%).
&
Progress increases with task advancement, remains unchanged during
repeated confirmation, and reaches 100\% only upon completion.
\\
\midrule

Failure recognition without recovery
&
Search (0\%)
$\rightarrow$ recognize that no result satisfies all constraints (0\%)
$\rightarrow$ temporarily overestimate progress (100\%)
$\rightarrow$ identify the selected item as unsuitable (0\%)
$\rightarrow$ remain in the unresolved failure state (0\%).
&
The estimate recovers from a temporary overestimation and ultimately
reflects failure, although the policy does not recover the task.
\\
\midrule

False-positive progress
&
Search (0\%)
$\rightarrow$ mistake a Henley T-shirt for a dress shirt (100\%)
$\rightarrow$ select an incorrect color (0\%)
$\rightarrow$ select size 2X and color navy (50\%)
$\rightarrow$ treat the Henley T-shirt as satisfying the request (100\%).
&
Misinterpreting a core product attribute produces overconfident
progress estimates and an incorrect purchase.
\\
\bottomrule
\end{tabularx}
\label{tab:qualitative_progress}
\end{table}

The successful case shows that progress estimates can reflect meaningful task transitions rather than increase mechanically with trajectory length: repeated confirmation actions retain the same progress value.
The second case separates progress assessment from action recovery. The agent eventually recognizes the failed state but cannot find a successful alternative.
The third case exposes a key limitation: progress assessment remains conditional on the agent's semantic understanding of the environment, so an incorrect interpretation can lead to confidently incorrect progress.

\section{Limitations}
RePro represents task progress through verbalized assessments,
making its estimates and rationales explicit; alternative
representations, such as latent states or hidden-vector supervision,
remain unexplored.
RePro also requires separate supervised fine-tuning and policy
optimization stages, adding training and pipeline overhead.
Future work could explore joint or interleaved training to
reduce this overhead while preserving stable progress assessment.
Our evaluation covers WebShop, ALFWorld, and Sokoban, while
resource and infrastructure constraints limit exploration of
broader settings such as operating system automation and open-ended embodied
interaction.
Finally, these benchmarks provide no ground-truth intermediate
progress labels. RePro's retrospective estimates use completed
trajectories and outcomes but may still misinterpret task states.
Future work could assess their calibration using human annotations
or stronger-model judgments, accounting for the uncertainty and
potential biases of these references.

\end{document}